\documentclass[]{spie}

\usepackage{amsmath,amsfonts,amssymb}
\usepackage{graphicx}
\usepackage[colorlinks=true, allcolors=blue]{hyperref}
\usepackage{algorithm}
\usepackage{algpseudocode}
\usepackage{booktabs}
\usepackage{multirow}
\usepackage{hyperref}

\title{AC-LoRA: Auto Component LoRA for Personalized Artistic Style Image Generation}

\author[a,b]{Zhipu CUI}
\author[b]{Andong Tian}
\author[b]{Zhi Ying}
\author[a]{Jialiang Lu*}
\affil[a]{SPEIT, Shanghai Jiaotong University, Shanghai, China}
\affil[b]{Ubisoft La Forge, Shanghai, China}
\authorinfo{*Corresponding Author}

\begin{document}

\pagestyle{empty}
\maketitle

\begin{figure}[ht]
    \centering
    \includegraphics[width=0.7\linewidth]{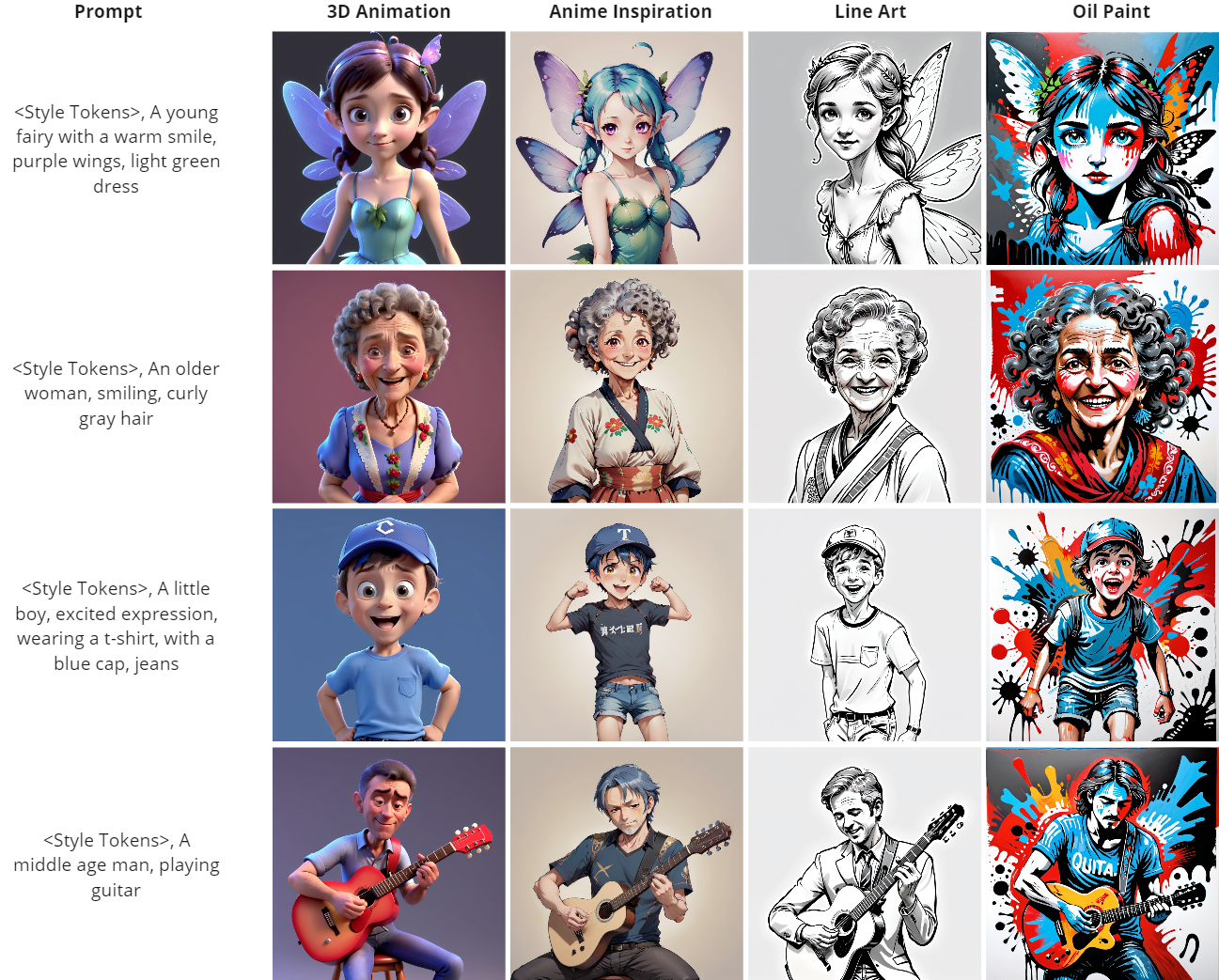}
    \caption{Personalized image generation results of four artistic styles based on four prompts using AC-LoRA.}
    \label{fig:teasor}
\end{figure}

\begin{abstract}
Personalized image generation allows users to preserve styles or subjects of a provided small set of images for further image generation. With the advancement in large text-to-image models, many techniques have been developed to efficiently fine-tune those models for personalization, such as Low Rank Adaptation (LoRA). However, LoRA-based methods often face the challenge of adjusting the rank parameter to achieve satisfactory results. To address this challenge, \textit{AutoComponent-LoRA} (AC-LoRA) is proposed, which is able to automatically separate the signal component and noise component of the LoRA matrices for fast and efficient personalized artistic style image generation. This method is based on Singular Value Decomposition (SVD) and dynamic heuristics to update the hyperparameters during training. Superior performance over existing methods in overcoming model underfitting or overfitting problems is demonstrated. The results were validated using FID, CLIP, DINO, and ImageReward, achieving an average of 9\% improvement.
\end{abstract}


\section{INTRODUCTION}
\label{sec:intro}

The task of transferring styles from one image to another is a long-standing computer vision problem and was previously considered as a problem of texture transfer \citenum{efros1999texture, jacobs2001image, lee2010directional}. The key is to apply the style from the source image while preserving the structure in the target image. With the development of deep learning techniques, neural network-based style transfer methods \citenum{zhu2017unpaired} outperform these non-parametric methods, showing promising results and enabling many applications.

Recent advances in generative models, and especially large text-to-image models, open the door to image generation. Users without artistic skills can create high-quality images and artwork by guiding the model with natural language \citenum{abdal2022clip2stylegan, andonian2021paint, gal2022stylegan, ojha2021few}. To better learn the data distribution for high-quality image generation, these models are trained on datasets containing millions or even billions of images \citenum{schuhmann2022laion}. However, in the case of personalized image generation in a specific artistic style, it remains challenging to control the style when directly guiding these pre-trained models with pure text prompts.

Thus, several personalized image generation methods for large text-to-image models have been developed \citenum{ruiz2023dreambooth}. These methods set a small number of images as the personalization target and attempt to adapt the large text-to-image model for personalized generation. The goal is to preserve the style or subject of the personalization target while leveraging the learned general image data distribution in the pre-trained base model. In particular, one of these solutions, Low-Rank Adaptation (LoRA) \citenum{hu2021lora, aghajanyan2020intrinsic, xu2022lora}, assumes that the matrix variations during training are low-rank and reduces the number of variable weights by introducing trainable low-rank matrices. However, this approach still suffers from a serious limitation: it introduces a new parameter, the rank of LoRA, which defines the dimension of the LoRA matrices and greatly influences the final result of the model. Depending on the personalization target data, a rank that is too low will lead to underfitting, while one that is too high will lead to overfitting. It is often difficult to find the optimal rank value without conducting multiple experiments.

To address the challenges in applying LoRA, a novel method called AutoComponent-LoRA (AC-LoRA) is proposed to automatically search for the best rank. This method allows high-quality and efficient training on very small datasets. Compared to other methods, AC-LoRA can significantly reduce the amount of time required for rank search. Theoretically, the technique can reduce the time required for model training of a given personalization target by an order of magnitude. At the same time, the final quality of the images generated by the model is improved compared to other methods since the algorithm automatically provides more accurate ranks.

The contributions of this work are as follows: 
\begin{itemize} 
    \item AC-LoRA is proposed to automatically search for the best rank based on SVD \citenum{zhang2015singular} eigenvalue analysis, addressing the challenge of finetuning large text-to-image models for personalized image generation. 
    \item The generalization of the algorithm is verified across different training datasets. To ensure this, the quality of the generated images of the AC-LoRA model trained on 8 datasets of 8 different art styles is validated separately. 
    \item The method is compared with other LoRA methods using various metrics. Results are validated with FID \citenum{heusel2017gans}, CLIP \citenum{hessel2021clipscore}, and DINO \citenum{caron2021emerging} scores. It is demonstrated that the quality of the images generated by the proposed model is higher than those of other LoRA models. 
\end{itemize}

\section{RELATED WORKS}
\label{sec:related_works}

\subsection{Personalized Image Generation}

Given a small set of existing images that contain the same subject or style, along with text prompt, the objective is to adapt text-to-image models in order to preserve the subject or style in generated images. There are two main categories of methods: finetuning-based methods and encoder-based methods.

\textit{Finetuning-based methods} follow the few-shot learning scheme. They use a very small number of target images to train the text-to-image model, which is based on a pre-trained base model. Textual Inversion \citenum{gal2022image} finetunes the text encoder to find new words in the textual embedding space, preserving a subject while modifying the context. DreamBooth \citenum{ruiz2023dreambooth} finetunes the pre-trained text-to-image model by embedding a given subject instance in the output domain and binding the subject to a unique identifier.

\textit{Encoder-based methods} \citenum{gal2023encoder} employ an additional image encoder to extract personalization target features and inject them into the diffusion model at attention layers. The encoder needs to be pre-trained on a large dataset. When performing inference, the user needs to provide the personalization target image.

\subsection{LoRA and Variations}
\label{sec:lora_variations}

LoRA is a method for efficiently fine-tuning large pre-trained models. It assumes that changes in weights during training are essentially low rank and splits the original weight matrix into the product of two much smaller matrices. For a pre-trained weight matrix $W_0 \in \mathbb{R}^{d \times k}$, LoRA constrains its update $\Delta W$ by representing it as a low-rank decomposition: $\Delta W = BA$, where $B \in \mathbb{R}^{d \times r}$, $A \in \mathbb{R}^{r \times k}$, and the rank $r \ll \min(d, k)$. Thus, the forward pass becomes:

\begin{equation}
y = W_{0}x + BAx
\end{equation}

\noindent where $y$ and $x$ represent the output and input, respectively. During training, $\Delta W$ is frozen, while $A$ and $B$ contain trainable parameters. This decomposition makes the fine-tuning process efficient and can adapt to small datasets.

Despite its advantages, LoRA still has some drawbacks. During fine-tuning, it introduces a new key parameter, rank $r$. The choice of this parameter can greatly affect the final performance of the personalized text-to-image model. Depending on the fine-tuning dataset, a rank that is too low will result in a model that does not have enough capability to represent the target personalization data, leading to underfitting. A rank that is too high will reduce efficiency, add significant noise into the weights, and result in overfitting, ultimately leading to degraded generated images. Therefore, selecting an appropriate rank for a specific target personalization dataset is crucial, which will require hyperparameter search with multiple experiments.

Dynamic Low-Rank Adaptation (DyLoRA) \citenum{valipour2022dylora} is a technique designed to enhance the adaptability of LoRA modules by allowing them to operate on a range of ranks rather than being fixed to a single rank. This approach enables a dynamic search for ranks by ordering the representations learned during training. However, computing on multiple ranks adds complexity. Furthermore, since a starting point and range for the search must be provided, this approach does not inherently reduce the number of hyperparameters, meaning that it still requires a large number of experiments to produce acceptable results.

Low-Rank Kronecker Product (LoKR) \citenum{edalati2022krona} optimizes training by decomposing large matrices into Kronecker products of multiple (normally 4) low-rank matrices, thereby significantly reducing the number of parameters and computational requirements. While this improves efficiency and reduces VRAM usage, similar to LoRA, it may affect the accuracy of the model by underfitting or overfitting on certain datasets. In addition, the large number of trainings required to select the optimal ranks of the matrices can also increase the complexity and time needed for the entire training process.

Other methods such as LoRA-FA \citenum{zhang2023lora-fa} and iA3 \citenum{houlsby2019parameter} also use various methods to improve LoRA. However, they face similar problems, with some variants producing unsatisfactory final results while others add too much computational stress.

\section{METHODOLOGY}
\label{sec:methodology}

\begin{figure}[h!]
    \centering
    \includegraphics[width=0.7\linewidth]{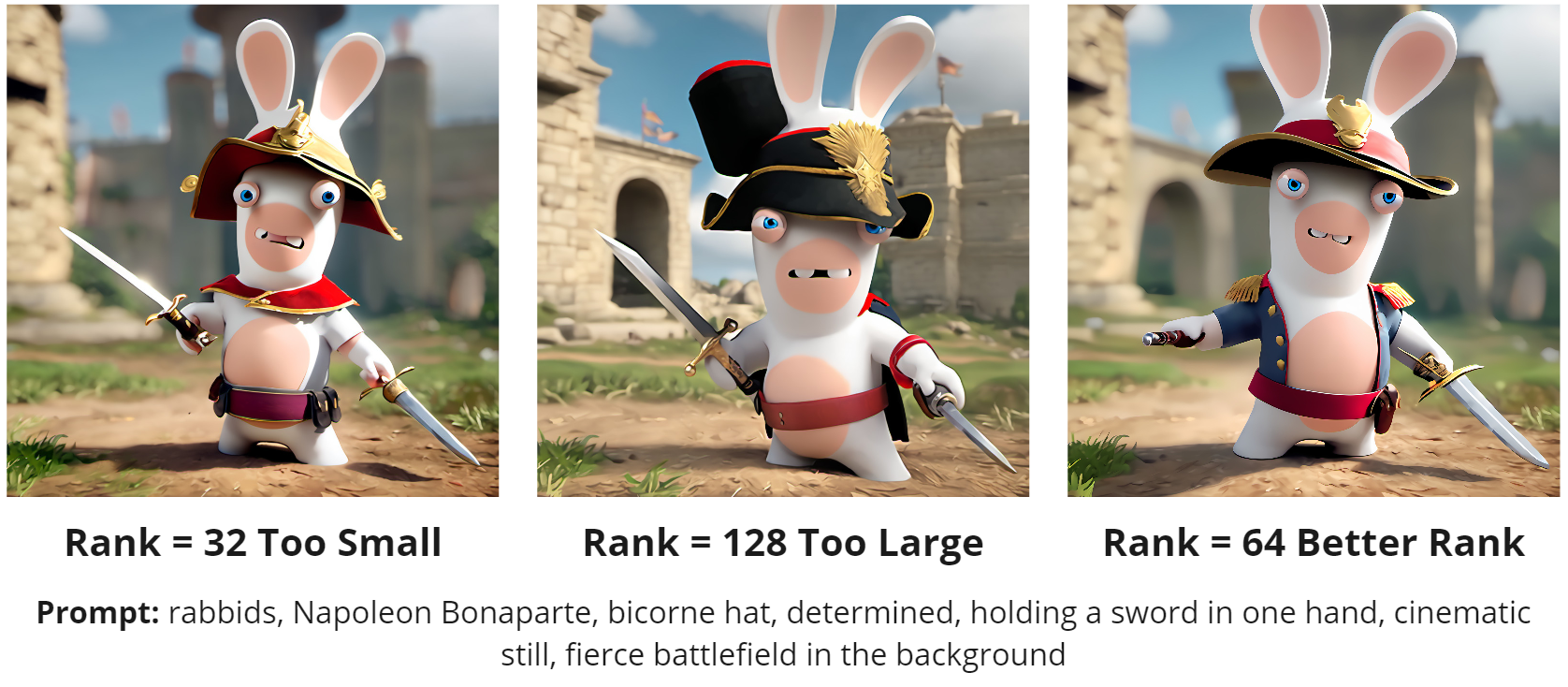}
    \caption{Comparison of generated images at different ranks}
    \label{fig:rankcompare}
\end{figure}

In this section, the focus is on the proposed AC-LoRA method, describing how the rank parameter of LoRA is automatically selected using the SVD technique.

A new LoRA variant, called AC-LoRA, is proposed to improve the original algorithm. The algorithm is designed to automatically search for the rank without adding any additional hyperparameters. This significantly reduces the number of experiments required to obtain a good model, thus reducing the training time for a given personalization target category by an order of magnitude or more. The method adjusts the LoRA matrix by adding corrections periodically during the original training process. The correction process identifies retained and discarded parts based on a threshold. The retained parts are left intact, while the discarded parts are transformed into Gaussian noise with the same variance. In this way, the new algorithm improves the overall training efficiency and provides better results due to the optimization of the rank.

\subsection{The Components of the LoRA Module}

Theoretically, the LoRA matrix contains three components: the signal component $M_S$, the noise component $M_N$, and the error component $M_{\epsilon}$. The signal component contains the main features of the dataset, which are the features the model should learn. The noise component contains the features that are exclusive to each piece of data in the dataset. These features are confusing, difficult to control, and not the desired ones, potentially causing the model to overfit. The error component is caused by the limited size of the model or insufficient training.

Even though each part cannot be strictly separated, it has been observed after learning on multiple personalization target categories that:
\begin{itemize} 
    \item Too low a rank gives poor results because it does not have enough capacity to learn the distribution of the training dataset; the common features of the dataset are corrupted, and therefore the model does not reproduce them well. This indicates that the model is underfitted in this case. 
    \item Too high a rank can also lead to poor results because the extra rank causes the model to fit over many exclusive features, which not only wastes VRAM but also introduces a lot of noise. In this case, the model is overfitted. 
\end{itemize}

As shown in Figure \ref{fig:rankcompare}, when the rank is 32 (too small), the model cannot correctly reproduce the main features of the Rabbids, resulting in errors in the depiction of the eyes and mouth. When the rank is 128 (too large), the model over-learns the characteristics of the Rabbids, resulting in issues with the depiction of the clothes. At a rank of 64, the model performs better.

\begin{figure*}[h!]
    \centering
    \includegraphics[width=0.7\linewidth]{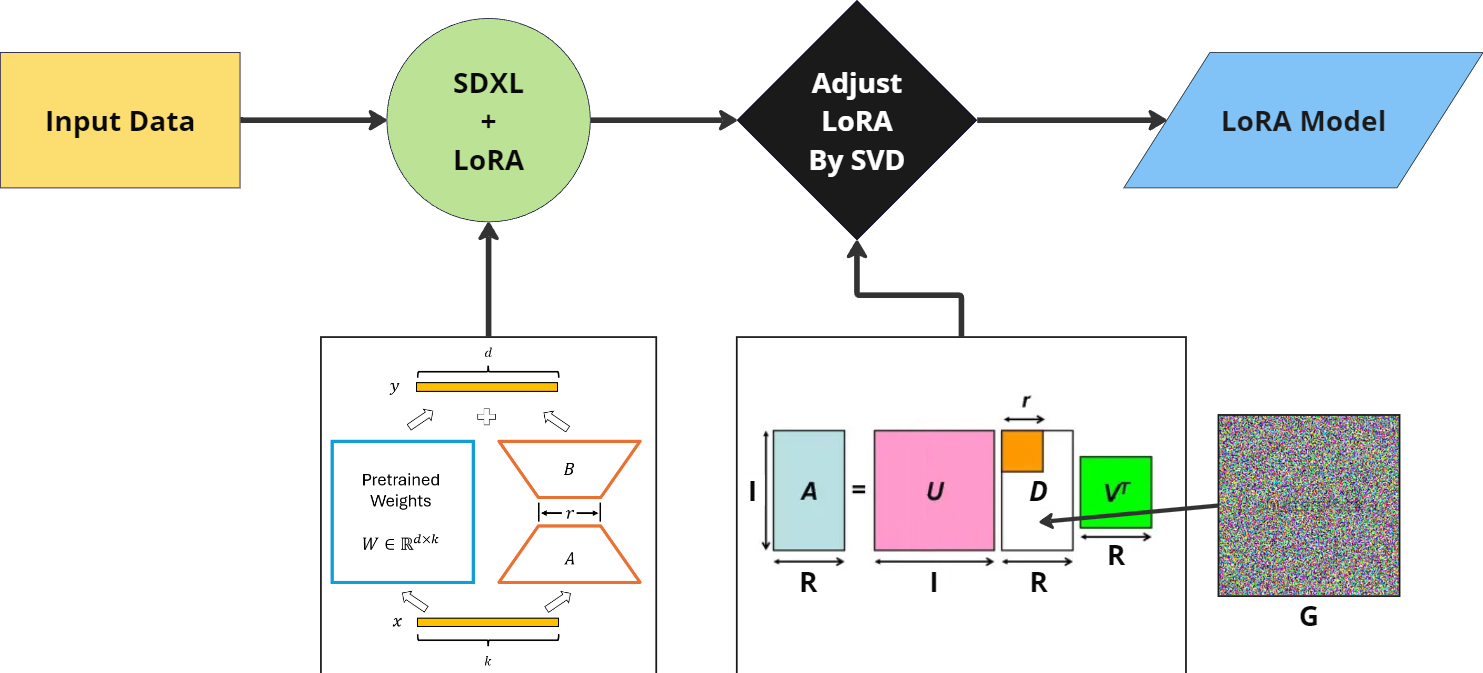}
    \caption{The overall pipeline of Auto Component}
    \label{fig:acpipe}
\end{figure*}

Upon further analysis, it can be argued that if a feature is one of the main features of the dataset, the model is bound to train this feature, and the weights will be upgraded in a fixed direction. The rank(s) in which this feature is located will, therefore, grow continuously, considering that LoRA is initialized with 0. On the other hand, if a feature is exclusive, the part of the rank(s) on which this feature is trained will vary in an unstable way, with no guarantee of continuous growth.

Based on the above phenomena and analysis, it is reasonable to assume that the ranks with high eigenvalues mainly contain the signal, whereas those with smaller eigenvalues primarily contain noise, which can be detrimental to the training process. Thus, removing these ranks is expected to help improve the quality of the final generated image.

\subsection{Auto Rank Search Method}

A method is proposed to automatically determine the rank based on SVD eigenvalue analysis. The LoRA matrix is decomposed and reorganized according to the following equations. This operation is defined as RESTART.

\begin{equation}\label{eq1}
\begin{split}
M &= UDV \\
D' &= \left\{\begin{matrix}
D_i & \text{if } i \in S \\
0 & \text{if } i \in N
\end{matrix} \right. \\
\sigma^2 &= \textrm{Var} \left ( UDV - UD'V \right ) \\
G &= G(0,\sigma^2) \\
M' &= UD'V + G
\end{split}
\end{equation}

\noindent where $M$ and $M'$ represent the LoRA matrix before and after RESTART, respectively. Here, $M$ is capable of representing both $A$ and $B$ of the LoRA module, thus $M \in \mathbb{M}{I \times R}$ or $M \in \mathbb{M}{R \times O}$, where $I$ and $O$ represent the input and output dimensions of the module. To avoid the VRAM peak being too high, a maximum value $R$ for the rank is set based on experience. $U$, $D$, and $V$ represent the matrix after SVD. $D'$ is the corresponding eigenvalue matrix after the RESTART operation. $D_i$ represents the original values of the corresponding rank in the matrix $D$. $G$ is Gaussian noise with a mean of 0 and the same standard deviation as the noise $M-UD'V$. It is defined on the ranks of the noise part and adapted to the size of $M$ by adding 0 to the matrix. By using this Gaussian noise matrix to replace the original matrix, the removal of information is ensured. In practice, considering that the absolute value in the signal part is far larger than that of the noise part, and that the signal part still contains a small amount of noise, $G$ is added to the entire matrix to further accelerate the calculation. An overall workflow is shown in Figure \ref{fig:acpipe}.

Based on the analysis above, the threshold is determined based on the matrix $D$, and the part of the squared progressive sum $Sum_i$ that is less than the squared sum $Sum$ multiplied by the percentage $p$ is selected as the signal part with signal indexes $S$, and the rest as the noise part with indexes $N$.

\begin{equation}\label{eq2}
\begin{split}
S &= \{i|Sum_i < Sum * p\} \\
N &= \{i|Sum_i \geq Sum * p\}
\end{split}
\end{equation}

The following reasoning can be drawn from the previously described analysis:

\begin{align}
M_L = M_S + M_N + M_{\epsilon}
\end{align}

\noindent where $M_L$ is the matrix of LoRA, and $M_S$ and $M_N$ are the matrices corresponding to the $S$ and $N$ parts as defined above. The $M_{\epsilon}$ represents the error of the model.

Therefore, the $M_N$ part is eliminated and replaced with pure Gaussian noise with the same variance. By this method, the unstable and highly biased information is transformed into null information while ensuring the norm of the matrix. In practice, a RESTART adjustment is performed every $E$ epochs to maintain the stability of the model.

\subsection{Choice of the Threshold}
With the above approach, the difficulty of rank selection is transformed into the selection of another parameter, namely the threshold $p$. To avoid adding additional hyperparameters and to shorten the training process for a given personalization target category, a method for automatically selecting the parameter $p$ is discussed.

Although the specific ratios of each LoRA matrix cannot be derived, the overall ratio, which is the loss $l$ of the training, is available. The parameter $p$ is defined as a function of the loss:

\begin{equation}\label{eq3}
\begin{split}
\frac{ \left |M_N + M_{\epsilon} \right |}{\left | M_L \right |} &= p_L \rightarrow p \\
p &= 1 - l^{\alpha}
\end{split}
\end{equation}

\noindent where $p_L$ is the real ratio of the LoRA layer $L$. Here, $\alpha$ represents the separation strength, which accounts for the changes in the model during the overall training process. According to the previous analysis, since $M_S$ and $M_N$ do not grow at the same speed and $M_S$ grows faster than $M_N$, the threshold should be increased accordingly. It is suggested to set $\alpha$ as follows:

\begin{align}
\alpha = \frac{Epoch}{TotalEpoch} + 1
\end{align}

According to the analysis, $p$ should increase along with training. Since $l < 1$, $\alpha$ should rise with $epoch$. Additionally, since $p$ should be $1 - l$ at the beginning of training, $\alpha$ must be greater than 1.

The original choice of rank is converted into a choice of threshold. The previous strategy of using the same rank for all layers is inherently flawed due to the presence of down-sampled and up-sampled layers in the base model, where the information density differs from layer to layer. Using the same threshold ensures that the rank for each LoRA layer is determined based on information rather than hyperparameters. This approach improves the model's performance while increasing efficiency.

\subsection{Multilayer LoRA}
A LoRA module can consist of two or more layers. For different layers within the same LoRA module, to ensure the completeness of the information, it is proposed to take the largest $S_j$ among them:

\begin{align}
S = \bigcup_j S_j
\end{align}

\noindent where $S_j$ represents the signal indexes of each layer.

With the above approach, $S$ and $N$ are defined and analyzed. The critical but difficult task of rank selection is transformed into a more accurate and automatic selection of $p$ using the SVD method. Finally, $p$ is selected based on $l$. These methods provide a complete automatic rank selection process without adding additional hyperparameters, greatly shortening the training process for a specific personalization target category. At the same time, the quality of the images generated by the model is improved due to more accurate rank selection and the use of different ranks in different layers. Additionally, since the RESTART operation is performed between two epochs, the model does not require higher VRAM and does not increase computational pressure.

\begin{algorithm}[H]
\caption{RESTART operation of the AC-LoRA}
\begin{algorithmic}[1]
    \Require Dataset, Base Model, LoRA Structure
    \Ensure LoRA Model
    \State Initialize: $l[]$ to store loss per epoch
    \For{$epoch = 0$ to $epochs$}
        \State Update model using LoRA
        \State Record loss $\rightarrow l[epoch]$
        \If{$epoch \% E == 0$}
            \State $L = avg(l)$
            \State $p = 1 - L^{\frac{epoch}{totalepoch - 1} + 1}$
            \ForAll{layers in LoRA Module}
                \State Perform SVD on weight matrix $\rightarrow U,D,V$
                \State Find max $i$: $\Sigma{D_i} < \Sigma{D} * p$
                \State $I = max(I,i)$
            \EndFor
            \ForAll{layers in LoRA Module}
                \State $D' = D$ with only first $I$ dims retained
                \State Record new weight $= UD'V$
                \State Record $\sigma[layer] = \text{std}(UDV - UD'V)$
            \EndFor
            \ForAll{layers in LoRA Module}
                \State Generate Gaussian $G$ with $\sigma[layer]$
                \State Update weights $\rightarrow UD'V + G$
            \EndFor
            \State Empty $l$
        \EndIf
    \EndFor
\end{algorithmic}
\end{algorithm}

\subsection{Theoretical Analysis}
Without loss of generality, consider $X_1$ and $X_2$, where $X_1$ takes random values on a unit hyper-sphere of dimension $R$, and $X_2$ is always constant on this hyper-sphere. From the basics of probability theory, it is known that:

\begin{align}
X_1 \sim \mathcal{N} \left ( 0, \frac{1}{R}I \right )
\end{align}

\noindent where $\mathcal{N}$ denotes a Gaussian distribution.

In this case, the following is obtained:

\begin{align}
\left | Y_1 \right |_2 \sim \sqrt{\frac{\lambda}{R}} \chi^{(R)}
\end{align}

\noindent where $\lambda$ is the number of samples and $Y_1$ is the sum of $X_1$.

Thus, the following is obtained:

\begin{align}
R_Y = \frac{\left | Y_1 \right |_2}{\left | Y_2 \right |_2} \sim \sqrt{\frac{1}{R\lambda}} \chi^{(R)}
\end{align}

\noindent where $Y_2$ is the sum of $X_2$.

It can be concluded that $R_Y \rightarrow 0$ as $R$ becomes large and $\lambda$ becomes large. This indicates a high probability that the last $1 - l^\alpha$ part is biased because the values are too small compared to the others. With high probability, this is due to severe instability in the direction of convergence during the training, suggesting that these parts do not represent common features or contain too many biased features.

\section{RESULTS AND EVALUATION}

In this section, to perform efficient and high-quality training, a set of data preparation procedures is first defined. Then, the basic setup and specific steps of the experiments are described. The proposed method is compared with other LoRA methods on 8 datasets to validate its effectiveness. Four evaluation methods—FID, CLIP, DINO, and ImageReview—are used to validate the effectiveness of the approach.

\subsection{Dataset Preparation}

Existing personalized image generation datasets, such as DreamBooth \citenum{ruiz2023dreambooth}, do not satisfy the requirements for artistic style control. Therefore, a new dataset called the AC-LoRA dataset has been organized, consisting of 8 personalization target categories. Each category contains a training and a test set, with image and caption pairs. Each category includes 15 high-quality images at a resolution of 1024x1024. The caption consists of a sentence or several expressions describing the main content of the image, starting with a class token and containing only the content to be learned by the model. The dataset can be accessed at: \url{https://anonymous.4open.science/r/AC-LoRA-Dataset-8C53}.

For example, a prompt could be: "rabbids, a rabbid, exhausted from its hike, stands atop a cliff of a towering mountain, equipped with an outdoor backpack and clutching hiking poles, with a bay nestled at the mountain's base, all under the hues of a dusky sky."

\begin{figure*}[htpb!]
    \centering
    \includegraphics[width=0.7\linewidth]{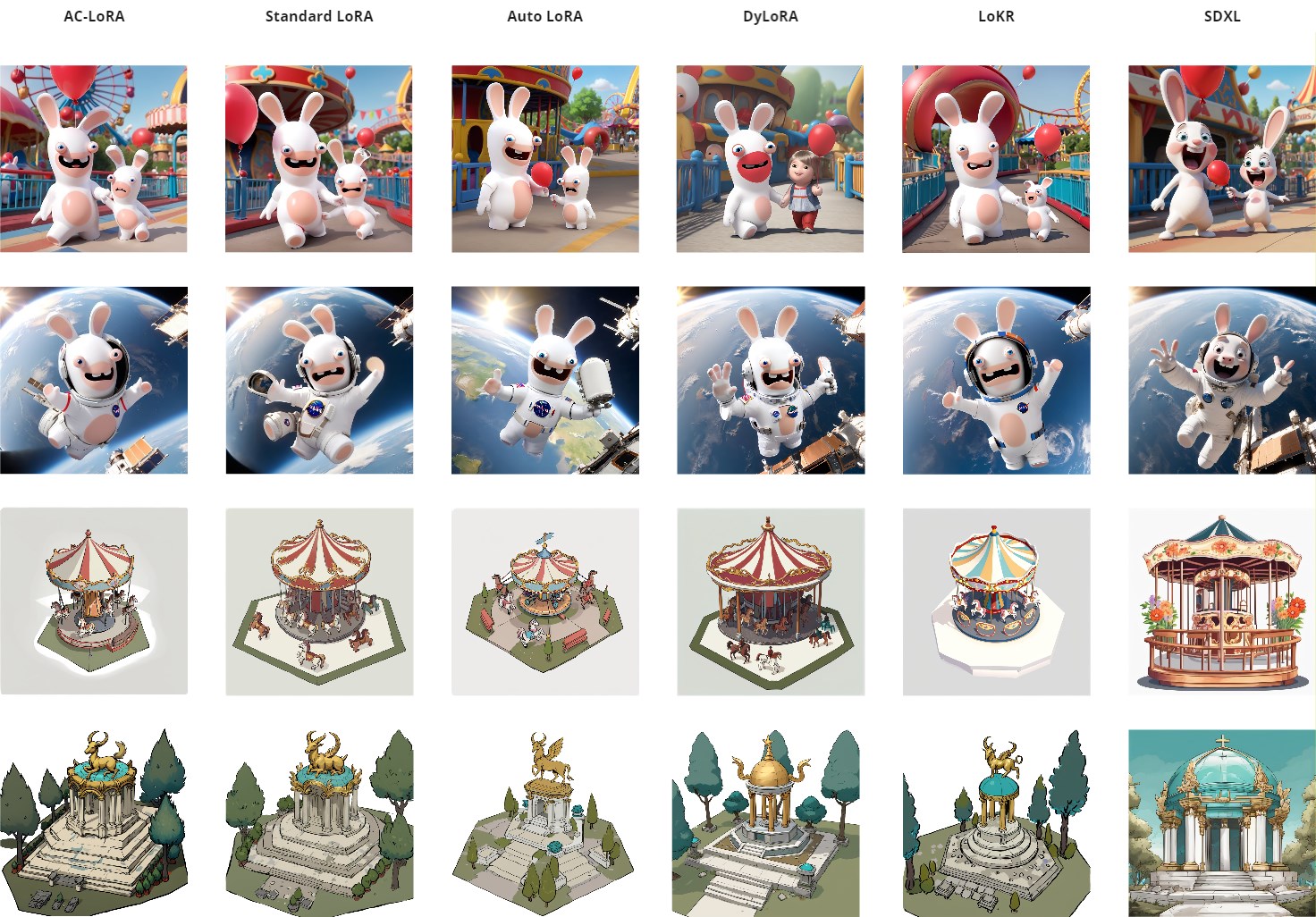}
    \caption{The comparison among AC-LoRA, other LoRAs, and the base model}
    \label{fig:loracompare}
\end{figure*}

\subsection{Training Environment and Cost}

Training was performed on hardware consisting of an Intel(R) Xeon W-2255 CPU and an NVIDIA GeForce RTX 3090 GPU, running Python 3.10.14. A total of 3000 different experiments were executed, each running for 100 epochs. The RESTART procedure was executed every 10 epochs. The VRAM requirement varied between 16.5 GB and 22.5 GB, with the peak not exceeding 22.5 GB. The duration was approximately 60 minutes per 100 epochs.

\begin{table*}[htpb!]
\centering
\scriptsize
  \caption{Evaluation results of AC-LoRA compared to other methods on 8 different topics}
  \label{tab:evaluaton}
  \begin{tabular}{l|ccc|ccc}
    \toprule
    \multirow{2}{*}{} &
      \multicolumn{3}{c|}{\textbf{AC-LoRA}} &
      \multicolumn{3}{c}{\textbf{LoRA}} \\
      & {FID $\downarrow$} & {CLIP $\uparrow$} & {DINO $\uparrow$} & {FID $\downarrow$} & {CLIP $\uparrow$} & {DINO $\uparrow$} \\
    \midrule
    Rabbids & \textbf{2.00} & \textbf{0.89} & \textbf{0.72} & 2.12 & 0.87 & 0.51 \\
    TheSettlers & \textbf{0.93} & \textbf{0.96} & \textbf{0.75} & 1.08 & 0.95 & 0.60 \\
    LineArt & \textbf{1.21} & \textbf{0.99} & \textbf{0.81} & 1.61 & 0.97 & 0.65 \\
    OilPaint & \textbf{0.74} & \textbf{0.96} & \textbf{0.87} & 1.70 & 0.96 & 0.78 \\
    3DAnimation & \textbf{1.14} & \textbf{0.99} & \textbf{0.83} & 1.23 & 0.99 & 0.72 \\
    AnimeInspiration & \textbf{1.29} & \textbf{0.90} & \textbf{0.82} & 1.31 & 0.89 & 0.79 \\
    VectorArt & \textbf{1.76} & \textbf{0.96} & \textbf{0.85} & 1.94 & 0.94 & 0.79 \\
    DigitalIllustration & \textbf{1.25} & \textbf{0.95} & \textbf{0.78} & 1.48 & 0.93 & 0.72 \\
    \midrule
    \multirow{2}{*}{} &
      \multicolumn{3}{c|}{\textbf{AutoLoRA}} &
      \multicolumn{3}{c}{\textbf{DyLoRA}} \\
      & {FID $\downarrow$} & {CLIP $\uparrow$} & {DINO $\uparrow$} & {FID $\downarrow$} & {CLIP $\uparrow$} & {DINO $\uparrow$} \\
    \midrule
    Rabbids & 2.04 & 0.88 & 0.54 & 2.27 & 0.85 & 0.30 \\
    TheSettlers & 0.95 & 0.95 & 0.71 & 1.68 & 0.93 & 0.45 \\
    LineArt & 1.37 & 0.98 & 0.70 & 1.77 & 0.96 & 0.62 \\
    OilPaint & 1.04 & 0.96 & 0.76 & 1.73 & 0.95 & 0.55 \\
    3DAnimation & 1.34 & 0.99 & 0.77 & 1.66 & 0.99 & 0.68 \\
    AnimeInspiration & 1.41 & 0.90 & 0.78 & 1.80 & 0.87 & 0.67 \\
    VectorArt & 1.77 & 0.93 & 0.84 & 2.35 & 0.91 & 0.68 \\
    DigitalIllustration & 1.34 & 0.93 & 0.65 & 2.06 & 0.90 & 0.57 \\
    \midrule
    \multirow{2}{*}{} &
      \multicolumn{3}{c|}{\textbf{LoKR}} &
      \multicolumn{3}{c}{\textbf{SDXL}} \\
      & {FID $\downarrow$} & {CLIP $\uparrow$} & {DINO $\uparrow$} & {FID $\downarrow$} & {CLIP $\uparrow$} & {DINO $\uparrow$} \\
    \midrule
    Rabbids & 2.18 & 0.86 & 0.33 & 3.42 & 0.85 & 0.29 \\
    TheSettlers & 1.24 & 0.94 & 0.46 & 2.61 & 0.92 & 0.43 \\
    LineArt & 2.05 & 0.96 & 0.58 & 2.23 & 0.96 & 0.53 \\
    OilPaint & 1.72 & 0.95 & 0.68 & 1.88 & 0.95 & 0.54 \\
    3DAnimation & 1.44 & 0.99 & 0.62 & 2.19 & 0.99 & 0.61 \\
    AnimeInspiration & 1.65 & 0.88 & 0.77 & 2.28 & 0.87 & 0.60 \\
    VectorArt & 2.13 & 0.91 & 0.71 & 2.81 & 0.91 & 0.67 \\
    DigitalIllustration & 1.81 & 0.89 & 0.55 & 2.37 & 0.89 & 0.49 \\
    \bottomrule
  \end{tabular}
\end{table*}

\subsection{Result Evaluation}

The experiments were conducted on 8 different personalization target categories, including 6 personalization target categories from public datasets and 2 personalization target categories from private datasets. Each dataset contains 15 images. The results of the fine-tuned model were compared with those of the base model, LoRA, and its variants—DyLoRA, LoKR, and AutoLoRA. The same strategy and dataset were used for training to control the variables.

Based on the examples shown in Figure \ref{fig:loracompare}, AC-LoRA outperforms the others both overall and in detail. The comparison of the 4 examples is as follows:
\begin{itemize} 
    \item AC-LoRA shows a distinct advantage over other LoRA methods in terms of ears and body shape, and the background is much clearer and more coherent. 
    \item The depiction of the Rabbids' hands is significantly better compared to other LoRA methods. 
    \item The distribution of horses on the carousel is closer to reality and more aesthetically pleasing, with no horses outside the structure. 
    \item The structure of the pavilion and the generation of the sculpture demonstrate greater advantages with AC-LoRA. 
\end{itemize}

Four evaluation techniques—FID, CLIP, DINO, and ImageReward—were used to assess the quality of the generated images. The results are shown in Table \ref{tab:evaluaton}.
\begin{itemize} 
    \item \textit{The Fréchet Inception Distance (FID)} is a popular metric for evaluating the quality of generated images by comparing feature distributions between generated and real images from the training dataset. FID scores, calculated using the Inception V3 network, indicate better quality with lower values. In the experiments, the approach achieved an average improvement of \textbf{2\% - 41\%} compared to the best of other methods. 
    \item \textit{The Contrastive Language-Image Pre-Training (CLIP)} method evaluates the alignment between an image and its text description using a similarity score derived from encoding both the image and text. Higher similarity scores indicate better quality of the generated output. In the experiments, the approach achieved an average improvement of \textbf{1\% - 3\%}. 
    \item \textit{The Distillation with No Labels (DINO)} method evaluates self-supervised learning models using a self-distillation mechanism with a teacher-student architecture. DINO measures the agreement between features extracted from both networks, with higher similarity scores indicating higher quality. In the experiments, the approach achieved an average improvement of \textbf{2\% - 34\%}. 
    \item \textit{ImageReward} evaluates the quality of generated images based on consistency with human preferences and criteria defined by a reward model. The model, trained on a dataset with human judgments, scores images on attributes like content accuracy, visual appeal, and adherence to prompts. It assigns a quantitative reward score to each image, with higher scores indicating greater consistency. In the experiments, the approach achieved an average improvement of \textbf{1\% - 7\%}. 
\end{itemize}

\section{Conclusion}

In this study, a new fine-tuning method for large text-to-image models, AC-LoRA, is proposed. Based on SVD matrix decomposition and dynamic heuristics, this method effectively addresses the challenge of selecting the optimal rank for LoRA. To better represent personalized artistic style image generation tasks, a new dataset was introduced. The method demonstrated superior performance in fine-tuning stable diffusion using this dataset compared to other LoRA variants. From the experimental results, the method successfully generated high-quality images for given artistic styles while leveraging the general image data distribution learned in the pre-trained base model.

A limitation of the method is that it still requires a certain amount of high-quality personalization target data (about 10 to 15 images) to achieve satisfactory results, a constraint inherited from the LoRA-based approach. In the future, exploring solutions such as pre-trained feature encoders to reduce this requirement will be considered. Another potential limitation is the model's robustness to noisy or adversarial inputs. Future research could focus on improving the model’s ability to handle noise and enhance robustness through adversarial training or noise-resistant techniques.

The approach for separating signal and noise components from LoRA matrices can also be applied to other tasks. It can serve as a general tool for feature reduction in probabilistic generative models. Further applications of this method are anticipated in the future, such as in domains like content creation, artistic image generation, and other personalized generative tasks.

\bibliography{report}
\bibliographystyle{spiebib}

\end{document}